%% file: elsarticle-template-num.tex
\documentclass[preprint,12pt]{elsarticle}

\usepackage{amssymb}
\usepackage{algorithm}
\usepackage{algorithmic}

\usepackage{amsmath}

\usepackage{xcolor}
\usepackage{listings}
\usepackage{pifont}

\usepackage{doi}
\usepackage{hyperref}
\hypersetup{
    hidelinks,  
}

\usepackage{booktabs}
\usepackage{graphicx} 

\usepackage{subcaption}

\journal{arXiv}

\begin{document}

\begin{frontmatter}



\title{LLMLogAnalyzer: A Clustering-Based Log Analysis Chatbot using Large Language Models}




\author[1]{Peng Cai\corref{cor1}}
\author[1]{Reza Ryan}
\author[1]{and Nickson M. Karie}
\affiliation[1]{organization={School of EECMS, Curtin University}, country={Australia}}

\cortext[cor1]{Corresponding author. Email: peng.cai@curtin.edu.au}

\begin{abstract}
System logs are a cornerstone of cybersecurity, supporting proactive breach prevention and post-incident investigations. However, analyzing vast amounts of diverse log data remains significantly challenging, as high costs, lack of in-house expertise, and time constraints make even basic analysis difficult for many organizations. This study introduces LLMLogAnalyzer, a clustering-based log analysis chatbot that leverages Large Language Models (LLMs) and Machine Learning (ML) algorithms to simplify and streamline log analysis processes. This innovative approach addresses key LLM limitations, including context window constraints and poor structured text handling capabilities, enabling more effective summarization, pattern extraction, and anomaly detection tasks. LLMLogAnalyzer is evaluated across four distinct domain logs and various tasks. Results demonstrate significant performance improvements over state-of-the-art LLM-based chatbots, including ChatGPT, ChatPDF, and NotebookLM, with consistent gains ranging from 39\% to 68\% across different tasks. The system also exhibits strong robustness, achieving a 93\% reduction in interquartile range (IQR) when using ROUGE-1 scores, indicating significantly lower result variability. The framework's effectiveness stems from its modular architecture comprising a router, log recognizer, log parser, and search tools. This design enhances LLM capabilities for structured text analysis while improving accuracy and robustness, making it a valuable resource for both cybersecurity experts and non-technical users.
\end{abstract}


\end{frontmatter}


\section{Introduction}
\label{sec:introduction}
Modern operating systems and applications produce a vast amount of log data at an incredible rate, making it a significant challenge to manage, analyze, and extract valuable information from log files in a timely and efficient manner. For the security of organizations or companies, it is critical to analyze log files to identify threats or responses to incidents. Security information and event management (SIEM) applications provide a variety of features to detect and prevent cyber-attacks, but most of them utilize predefined rules to identify anomaly behaviors \cite{González2021}. Traditional Machine Learning approaches require training a model with a training dataset and then applying the trained model to produce predictions \cite{Abdalla2022}. Although they work well in some situations, the training data set needs to be manually labeled by domain experts, the training process is time-consuming, and the model's quality depends on the quality of the training data set \cite{Chen2021}. Moreover, the predictions produced lack explanations. 
Egersdoerfer et al. explored the ability of LLMs in log analysis with an iterative refinement approach \cite{Egersdoerfer2023}, but their method is limited by the context capacity of the LLM, making it ineffective for analyzing large log files. Chen et al. introduced three ways to summarize large text files with LLMs, including map-reducing and iterative refinement \cite{Chen2024}. However, these approaches are computationally intensive and time-consuming.
To address the above challenges, this study introduces LLMLogAnalyzer, a clustering-based log analysis application that enables non-technical users to effectively summarize and analyze log files through conversations. It leverages the power of ML techniques and LLMs to provide a comprehensive log analysis solution, featuring keyword search, event search, semantic search, clustering-based summary, and analysis, making log analysis more accessible and user-friendly. 

\section{Background}
\label{sec:background}
\textbf{Log analysis} is the process of extracting insights from large volumes of log data, including identifying anomalous behavior and optimizing system performance \cite{Debnath2018}. As noted by He et al., log analysis typically involves two main stages: log parsing and log mining \cite{He2016}.

\textbf{Log parsing} is the initial step in log analysis, where unstructured log data is converted into structured events \cite{MaLLMParser2024}. Specifically, log parsing uses a clustering algorithm to extract static log templates and dynamic variables from the raw log file. Those logs that share the same log template can be considered as a type of log event. 

\textbf{Log-based anomaly detection}, a key task in log mining, involves identifying data points that exhibit unusual behavior or patterns that differ from the norm, allowing the detection of irregular behavior in log files \cite{Ahmed2016, Aussel2018}. 

\textbf{Large language models (LLMs)} are machine learning models, typically based on the Transformer architecture with hundreds of billions of parameters, pre-trained on large-scale data sets \cite{Chang2024survey}. LLMs have shown remarkable performance on various natural language processing (NLP) tasks \cite{Chang2024survey}.

\textbf{Retrieval Augmented Generation (RAG)} is an architecture that mitigates LLM hallucination problems and expands their knowledge sets by incorporating external sources without re-training LLMs \cite{Li2025retrieval}. As new information becomes available, pre-trained models may not be able to answer all questions. However, RAG provides a cost-effective way to keep LLMs up-to-date with the latest information, thereby avoiding the need for costly and time-consuming retraining of the models \cite{Procko2024graph}. A basic RAG architecture consists of three stages: indexing, retrieval, and generation, allowing efficient similarity searches in a high-dimensional space, as described by \cite{Li2025retrieval}. By retrieving relevant information from external documents and combining it with the user's query, the model can generate more accurate responses.

\section{Related Work}
\label{sec:relatedWork}
\subsection{Log Analysis  Development}
In the early days, administrators manually analyzed log files to identify errors and issues. However, as log files grew in size and complexity, log management tools emerged to help manage and analyze the data. Examples of such tools include Splunk, ELK (Elasticsearch, Logstash, Kibana) Stack, and Wazuh. 
As the log files continued to evolve, ML models were trained and applied to identify anomalies and predictive analytics. Machine learning-based log parsing algorithms can be broadly categorized into three primary approaches: density-based log parsing, frequent pattern mining, and heuristic approaches \cite {Zhang2023}. Density-based clustering algorithms, such as the Density-Based Spatial Clustering (DBSCAN) \cite{ester1996density}, group log messages into clusters based on density and distribution. Frequent pattern mining, such as SLCT (Simple Logfile Clustering Tool) \cite{vaarandi2003data}, discovers recurring patterns in log messages to identify common system events, error messages, or usage trends. Heuristic approaches, such as Drain \cite{He2017}, leverage logical rules and expertise to parse logs effectively. Drain, in particular, employs a fixed-depth parsing strategy utilizing a Parsing Tree data structure. 
The application of LLMs to log parsing has gained significant attention in recent research, with several studies showcasing improved results compared to traditional methods. Notable examples include LogPPT \cite{Le2023}, Llmparser \cite{MaLLMParser2024}, LILAC\cite{Jiang2024}, and DivLog \cite{Xu2024}, which state-of-the-art performance in log parsing tasks.

\subsection{LLMs for Log Analysis}
Apart from log parsing, large language models have been applied to anomaly detection in log files, with studies exploring the use of GPT-2 \cite{Han2023}, GPT-3.5 Turbo \cite{Qi2023}, and fine-tuned GPT-3 \cite{Hadadi2024}, thereby contributing to the growing body of research in this area.

Moreover, the introduction of PreLog \cite{Le2024}, a pre-trained model designed for log parsing and anomaly detection, has demonstrated its effectiveness through prompt tuning. In addition, a benchmarking framework for LLMs has been proposed, covering a range of activities such as log parsing, log-based anomaly detection, fault diagnosis, and summarization \cite{Cui2024}. Furthermore, the use of LLMs in online log analysis with advanced prompt strategies has been explored, suggesting opportunities for advancement in software operations and maintenance, as well as program comprehension \cite{Liu2024}. However, the development of a Question Answering (QA) pipeline using RAG has made a significant contribution to log-based anomaly detection, demonstrating promising performance \cite{Pan2024}.

Despite the advancements in log analysis, a notable gap exists in the development of LLM-based chatbots. This limitation makes log analysis challenging and underscores the need for innovative solutions. To bridge this gap, this study proposes LLMLogAnalyzer, an LLM-based chatbot tailored for log analysis. 
LLMLogAnalyzer leverages the strengths of ML and LLMs to enhance log analysis accuracy, efficiency, and effectiveness. This comprehensive solution covers various log analysis tasks such as summarization, pattern extraction, log-based anomaly detection, root cause analysis, etc. Moreover, LLMLogAnalyzer supports diverse log types, ensuring adaptability across various domains and systems such as Linux, Windows, and macOS.


\section{Proposed approach}
\label{sec:proposedApproach}
The proposed approach, LLMLogAnalyzer, integrates the strengths of ML and LLM to enhance log file analysis. By employing the clustering algorithm Drain with default benchmark settings for log parsing, the approach converts raw log data into structured data, facilitating the efficient detection of clusters of related events. Clustering algorithm Drain was selected for its exceptional parsing accuracy performance. When users send a query, LLMLogAnalyzer analyzes the user's query and directs it to the proper stage. For example, this may involve selecting suitable search tools to extract relevant information from structured log data, followed by generating responses based on the user's query and the search result. The responses include relevant references from the search results, providing a comprehensive and informed answer to the user's query. 

\subsection{Framework}

\begin{figure}[H]
    \centering
    \includegraphics[width=1.0\linewidth]{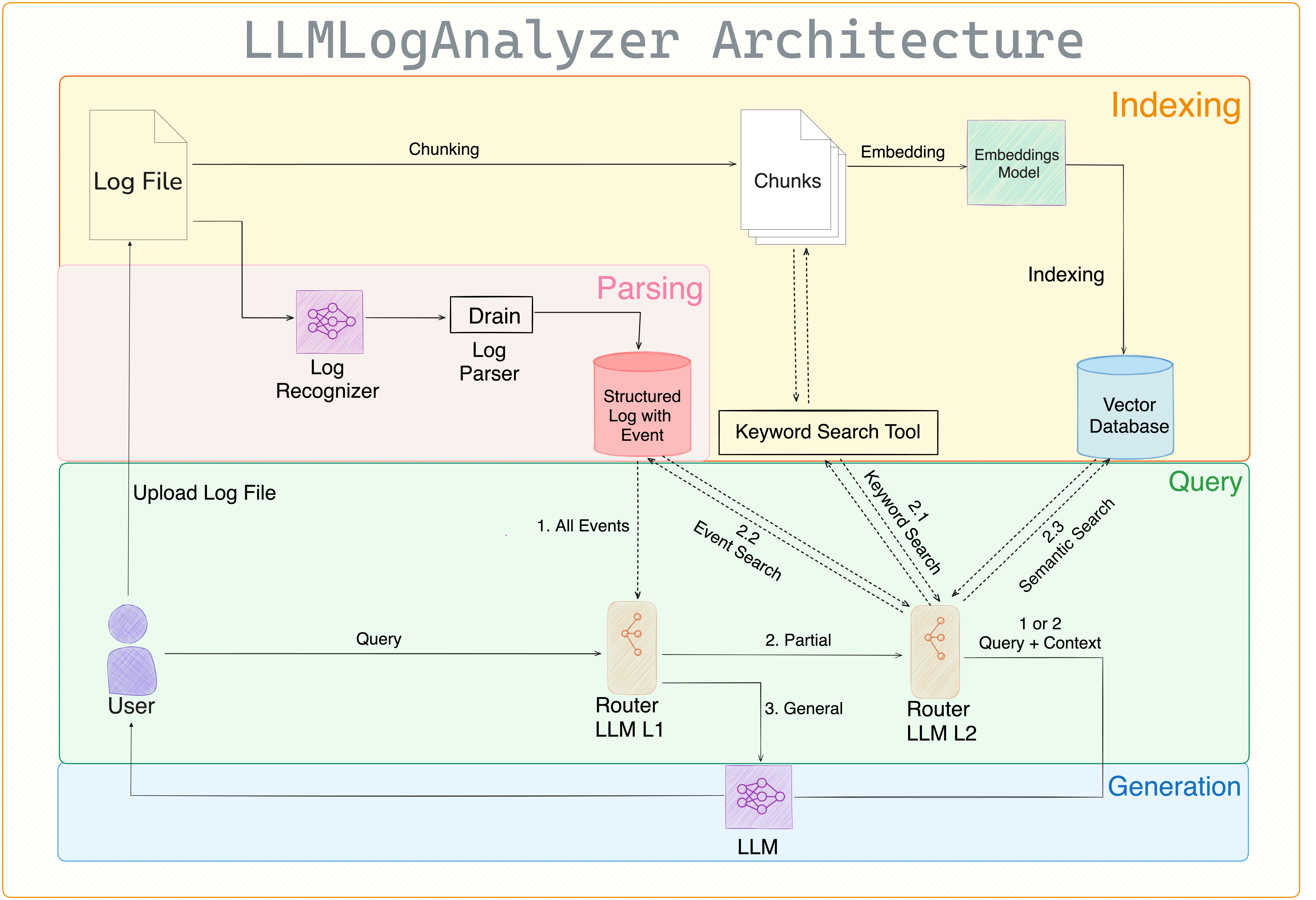}
    \caption{LLMLogAnalyzer Architecture}
    \label{fig:framework}
\end{figure}

The LLMLogAnalyzer framework presents a systematic methodology for log analysis that consists of four sequential stages: indexing, parsing, query, and generation. This framework is supported by four key components: routers, log recognizer, log parser, and search tools. The integration of the Drain algorithm and RAG architecture enhances the log analysis capabilities. Its modular architecture facilitates efficient processing and analysis across diverse log categories, leveraging flexible integration and scalability. Figure \ref{fig:framework} shows the high-level LLMLogAnalyzer architecture, showcasing how these components and stages work together to support comprehensive log analysis.

\input{algorithm}

The LLMLogAnalyzer algorithm is a four-stage process that analyzes log files and generates responses to user queries, as outlined in Algorithm \ref{alg:llmloganalyzer}. The indexing and parsing stage is performed when a log file is uploaded to LLMLogAnalyzer. In contrast, the query and generation stages are triggered dynamically in response to user queries during chat sessions, enabling on-demand log analysis and insightful responses. The same system prompt template is used consistently across all generation stages, with details provided in \ref{subsec:system_prompt_template}.

To support semantic search via the RAG architecture, the indexing stage involves dividing raw log files into multiple chunks and transforming tokens into vector representations using an embedding model. The resulting vector embeddings are then stored in a vector database, facilitating efficient semantic search and retrieval of relevant log information. 

LLMLogAnalyzer's parsing stage employs LLM for automatic log type identification. Once the log type is accurately recognized, the Drain algorithm is applied to parse raw log files into a structured format, associating each log entry with a specific event. After completing these preprocessing steps, LLMLogAnalyzer is ready for interactive log querying and analysis.

During the query stage, LLMLogAnalyzer initiates query analysis to discern the user's intent, thereby facilitating precise log retrieval and insightful analysis. The Router, a critical component of LLMLogAnalyzer, performs query analysis, categorizing queries into three tiers based on the required log context.

In cases where exhaustive log context is essential, the router directs queries to the All Event tier. This tier involves analyzing the entire structured log data to identify relevant patterns and relationships, providing definitive answers to complex queries.

Queries requiring specific log entities or segments are routed to the Partial tier for precise analysis. This tier employs three specialized tools: a keywords search tool that searches raw log data using relevant keywords to identify matching entries, an event search tool that queries structured log data using event IDs to retrieve specific log segments, and a semantic search tool that utilizes semantic analysis to retrieve the top two most similar chunks based on similarity with the user's query.

The General tier serves as a context-free processing level where queries that can be answered independently of log data are directed. This tier leverages the LLM's inherent knowledge and reasoning capabilities to respond, eliminating the need for log context analysis.

In the final stage, LLMLogAnalyzer generates a response to the user's query, leveraging context from the three tiers and the user's query.

Based on this approach, LLMLogAnalyzer is able to work as a chatbot and answer user queries related to log files, providing a promising solution for performing log analysis without log expertise. The following subsections elaborate on each stage of the framework.

\subsection{Indexing}
\begin{figure}[H]
    \centering
    \includegraphics[width=0.8\linewidth]{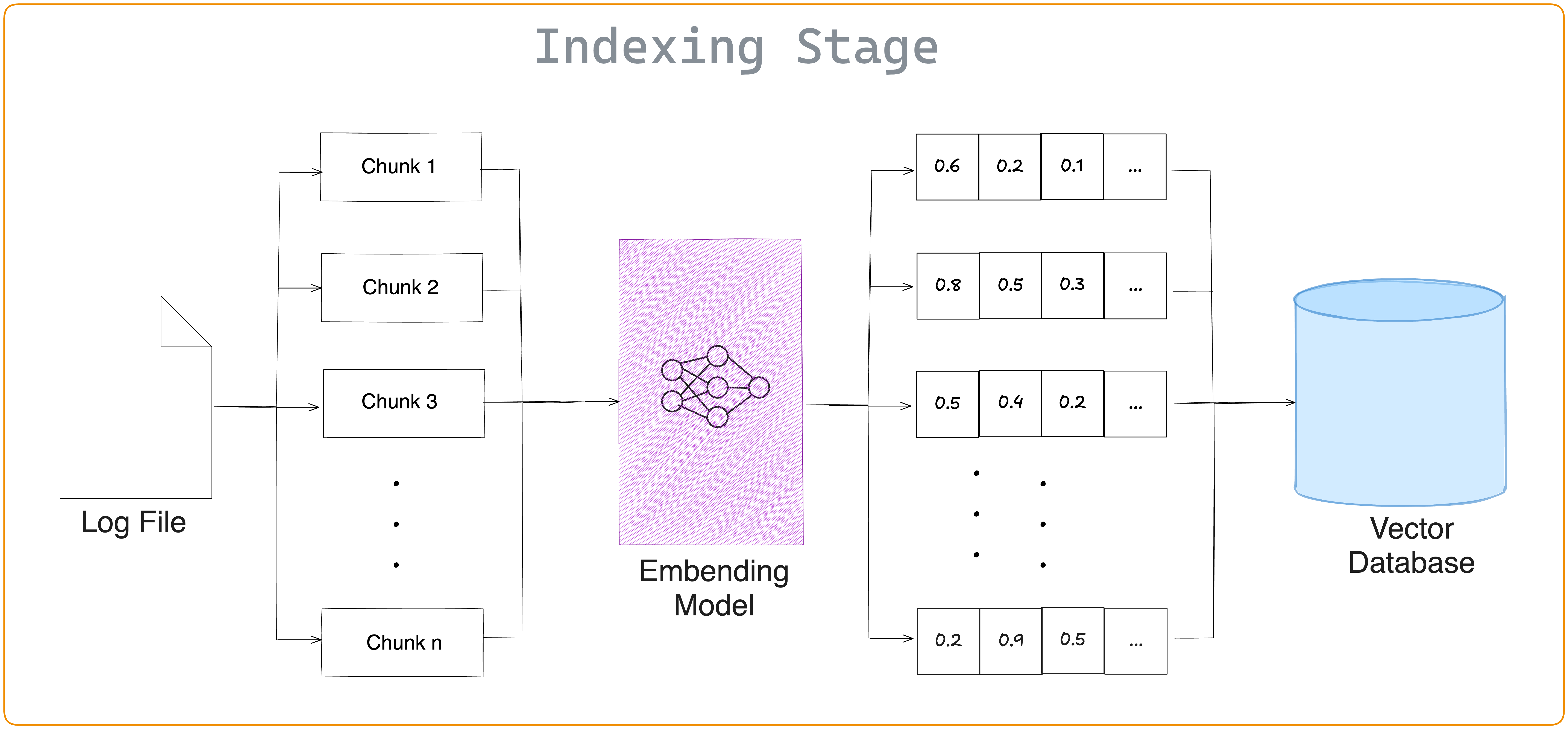}
    \caption{Indexing Stage}
    \label{fig:indexing}
\end{figure}

This stage involves creating an index of the log data to enable efficient semantic searching via RAG architecture. The indexing process consists of two crucial steps. First, the raw log file is chunked into smaller segments, typically 1024 tokens per chunk, to enable parallel processing and optimize computational efficiency. Next, an embedding model is applied to each chunk, converting the text data into vector representations, as shown in Figure \ref{fig:indexing}. These vectors are then stored in a vector database using LlamaIndex, which facilitates quick searching and retrieval.

\subsection{Parsing}
LLMLogAnalyzer begins by using LLM to identify the log type from the raw log file. This log type is then used as a parameter for the Drain algorithm. The Drain algorithm is then used to parse raw logs into a structured format with associated events.

Two key framework components are employed in this stage: the log recognizer and the log parser. Specifically, log recognizer leverages the capabilities of LLM to identify log categories. Log parser employs Drain to transform unstructured logs into a structured format with unique event IDs. These components work synergistically to preprocess and classify log data, laying the groundwork for subsequent analysis.

\subsubsection{Log Recognizer}
The log recognizer uses an LLM to analyze sample log data and identify the log type from among 16 supported candidates, including Windows, Mac, Linux, and various application and system logs from Loghub. By employing an LLM, the log recognizer avoids reliance on regex patterns, enabling more accurate and flexible log type identification. The prompt template employed by the log recognizer is detailed in \ref{subsec:log_recognizer_prompt_template}.

\subsubsection{Log Parser}
A log parser was developed using the clustering algorithm Drain, which was selected due to its superior accuracy and effectiveness compared to other available algorithms and techniques. After the log recognizer identifies the log categories, the relevant parameters are passed to Drain to parse the raw log data. Each individual log message from the dataset is processed and grouped with other log messages that share similar structural patterns. Through this clustering process, Drain identifies and extracts unique log templates along with their corresponding variable parameters, as shown in Figure 3. This clustering transforms the raw, parsing log data into structured templates that can be more easily analyzed and processed by downstream applications. Drain groups of logs using a fixed-depth parse tree, where each level corresponds to a token position in the log message. It compares incoming log messages against existing log templates using a simple token-based similarity heuristic \cite{He2017}. The implementation maintains consistent configuration settings with the Drain benchmark, including similarity threshold, parse tree depth, and other parameters to ensure reproducible and comparable parsing performance across different log datasets. The Drain clustering evaluation follows the LogHub2.0 methodology and achieves high performance, with 0.99 accuracy on the 10-million-message HDFS dataset, indicating strong clustering quality\cite{He2017}.

\begin{figure}[H]
    \centering
    \includegraphics[width=0.8\linewidth]{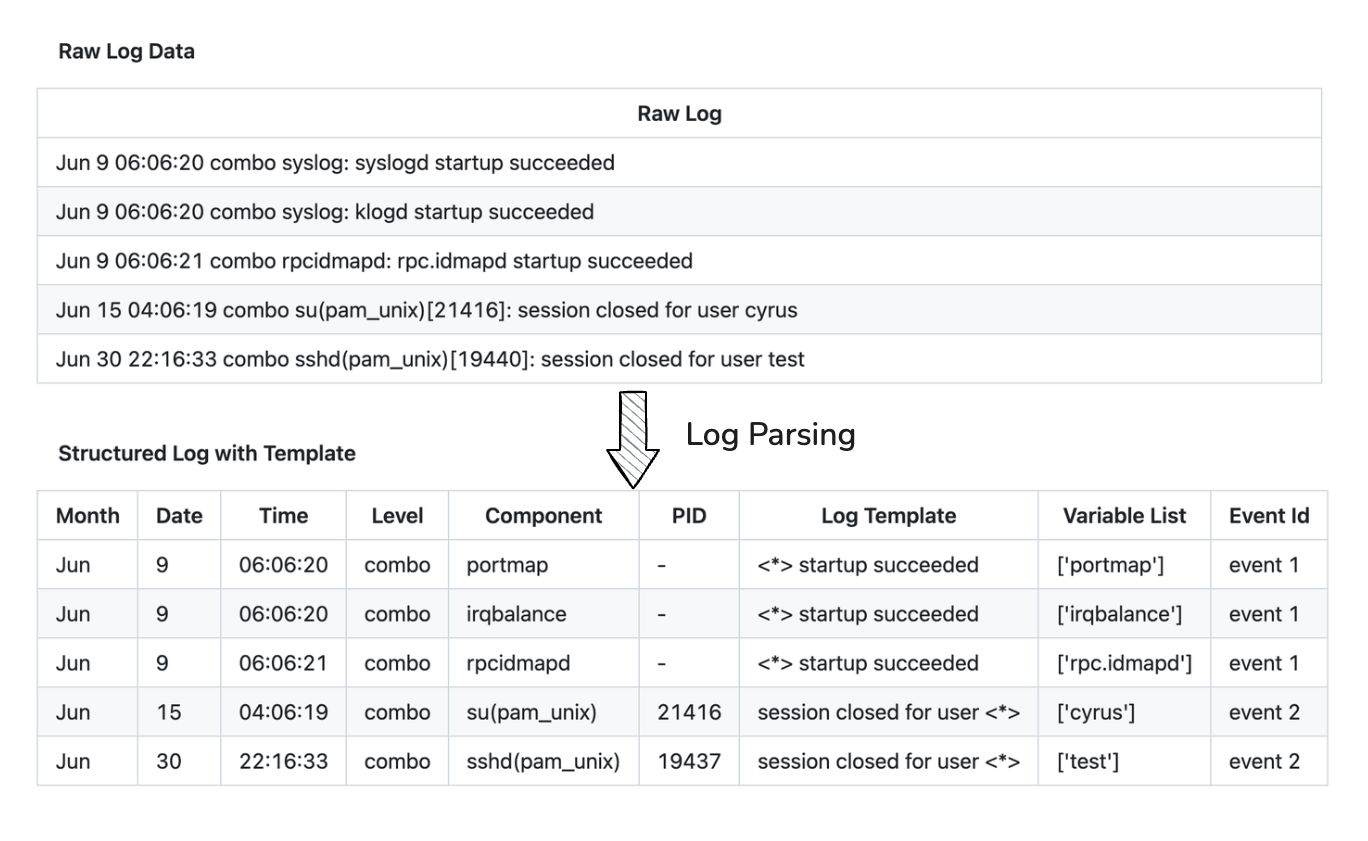}
    \caption{Parsing Stage}
    \label{fig:parsing}
\end{figure}

The parsing outcome is illustrated in Figure \ref{fig:parsing}, showcasing the transformation from unstructured to structured log data
Within the context of this research project, the log parsing process encompasses sixteen distinct categories.

\subsection{Query}

The LLMLogAnalyzer framework leverages the LLM's reasoning abilities in real time to dynamically select the most effective search tools and techniques based on user query analysis. By semantically analyzing user queries and routing them to appropriate processing stages, LLMLogAnalyzer delivers highly accurate and contextually relevant responses, improving the overall user experience.

The heart of efficient log analysis with LLMLogAnalyzer is the query stage, which consists of two key components: routers and search tools. The routers utilize LLM to analyze user queries and direct them to the appropriate processing stage via a two-level mechanism. This mechanism includes three level 1 options - all event, partial, and general - with level 2 providing additional specificity for partial queries through keyword, event, and semantic options.

\begin{figure}[H]
    \centering
    \includegraphics[width=0.8\linewidth]{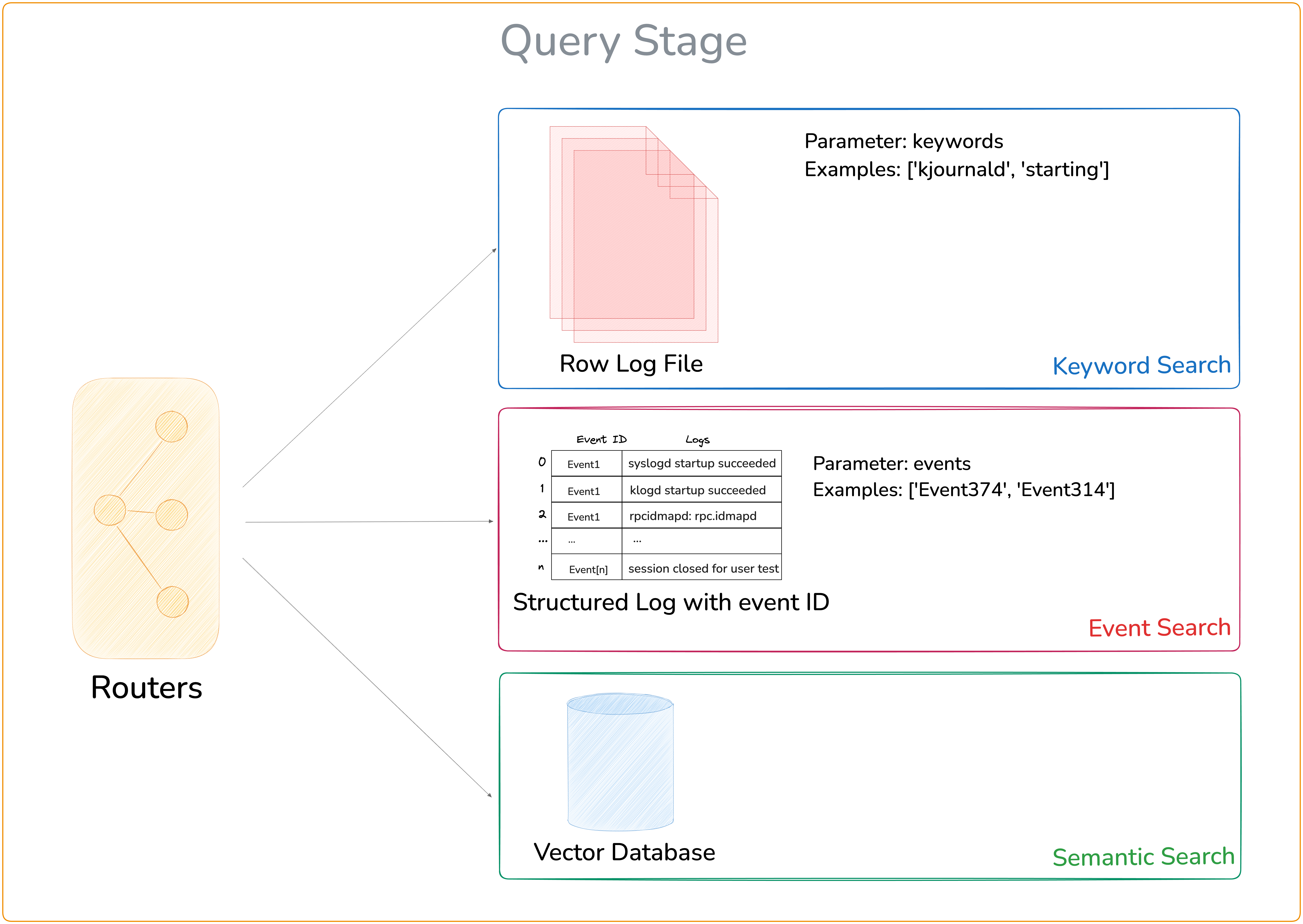}
    \caption{Query Stage}
    \label{fig:query}
\end{figure}

LLMLogAnalyzer supports three primary search tools: keyword search, event search, and semantic search, as illustrated in Figure \ref{fig:query}. These tools enable LLM to search log files through keyword matching, event ID filtering, and semantic retrieval of user queries. Specifically, the LLM generates a list of keywords for keyword searches and a list of event IDs for event searches. For semantic searches, the framework leverages LlamaIndex's advanced vector query engine, enabling sophisticated contextually relevant retrieval. The search tools are the powerhouse of the framework, providing comprehensive search functionality. 

Once the optimal search tool is selected, LLMLogAnalyzer invokes it with the provided parameters, retrieving results from the raw log file or structured log data. This seamless integration of search functionality enables comprehensive log analysis.

\subsubsection{Routers}

To efficiently direct user queries and optimise the performance of LLM in log analysis, a specialised routing system is implemented. The primary function of the router is to analyse user queries and determine the most effective way to retrieve relevant information from log files, striking a balance between computational cost and accuracy.

\begin{figure}[H]
    \centering
    \includegraphics[width=0.8\linewidth]{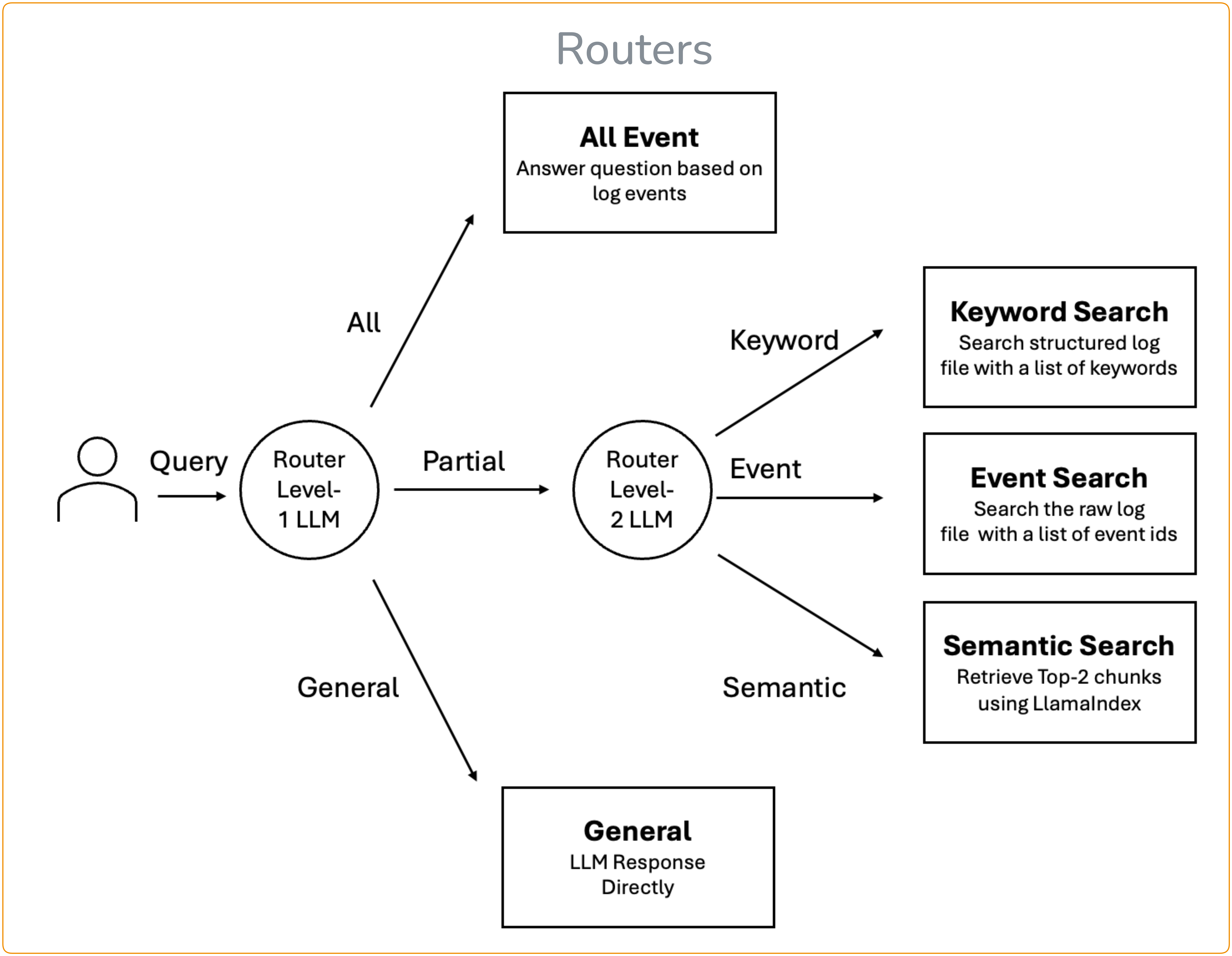}
    \caption{Routers}
    \label{fig:routers}
\end{figure}

Building on the query-based router achieved by Ding et al. \cite{Chen2024routerdc}, who successfully trained an LLM to act as a router, achieving an optimal balance between cost and efficiency, this research advances the state-of-the-art. The findings indicate significant potential for task-aware routers to improve LLM effectiveness in log analysis, motivating the creation of LLMLogAnalyzer. This contribution includes the design of a two-level routing system to optimize query processing, as shown in Figure \ref{fig:routers}

At the first routing level, user queries are classified into three distinct categories: all events, partial, and general, based on the LLM's query analysis. These categories dictate the extent of log file access required to provide context for LLM answering the user's query: "all event" queries require full structured events file access, "partial" queries require access to specific log segments or rows, and "general" queries do not require log data. The prompt template for router level 1 is shown in \ref{subsec:router_level_1_prompt_template}.

The second routing level presents three refined options - keyword, event, and semantic - exclusively when partial is selected, enabling precise data retrieval pathways. Keyword facilitates keyword matching, event supports event searching via event IDs, and semantic leverages RAG for semantic searches. The prompt template for router level 2 is shown in \ref{subsec:router_level_2_prompt_template}.

\subsubsection{Search Tools}

LLMLogAnalyzer offers three specialized tools for the efficient exploration of log data: keyword search, event search, and semantic search tools. Each tool provides distinct functionalities tailored to address specific analytical needs. LLMLogAnalyzer utilizes the best-suited tool selection to search relevant sources based on the user's query. This provides a comprehensive search capability for log data.

The keyword search tool, integrated into LLMLogAnalyzer, performs automated searches for specific keywords within log data. This enables efficient identification of relevant log entries, facilitates troubleshooting and analysis of log data, and quickly identifies error messages or exceptions related to a specific issue. In detail, the tool searches the log file for a list of keywords provided by LLM, checking each log entry for matching keywords with case insensitivity and returning matching entries. The result will be attached to the LLM response for easy reference.

\begin{figure}[H]
    \centering
    \includegraphics[width=0.8\linewidth]{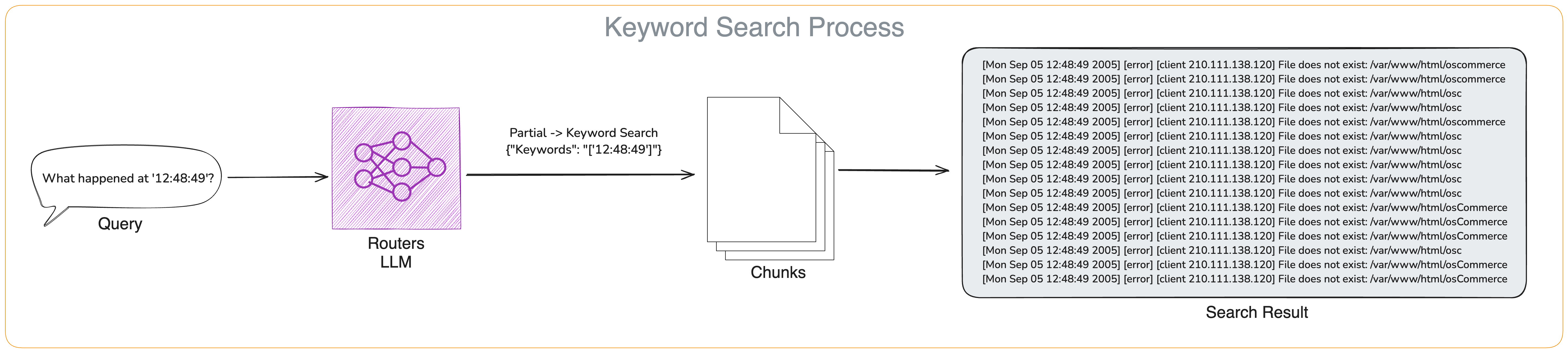}
    \caption{Keyword Search Process}
    \label{fig:keyword-search}
\end{figure}

As shown in Figure \ref{fig:keyword-search}, a keyword list, specifically ['03:28:22'], is provided by the LLMLogAnalyzer and then utilized for searching. The keyword search returns 8 matching log entities, which are then returned as part of the final response.

The event search tool, integrated into LLMLogAnalyzer, is designed to search the structured log files generated by Drain. These log files contain organized and formatted data, including event IDs, which enables the event search tool to accurately and efficiently locate matching events. Specifically, the tool performs automated searches for specific events, such as errors or warnings, to aid in troubleshooting and issue resolution. It searches for a specific list of events provided by LLM, checking the event IDs to identify matching log entities. The tool streamlines the analysis process, enabling users to promptly address issues and optimize system performance.

\begin{figure}[H]
    \centering
    \includegraphics[width=0.8\linewidth]{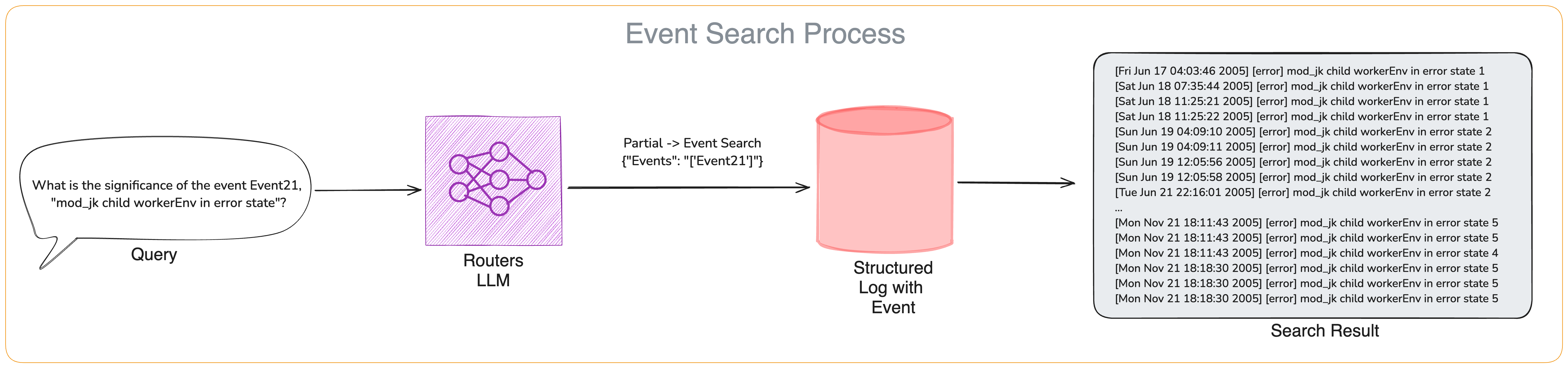}
    \caption{Event Search Process}
    \label{fig:event-search}
\end{figure}

Figure \ref{fig:event-search} illustrates the process where the LLMLogAnalyzer generates an event list, such as ['Event21'], which is subsequently used for the search operation. This query identifies 1,462 matching log entities, but the result has been truncated for illustrative purposes.

The semantic search tool is implemented using RAG architecture, leveraging LlamaIndex. When a user's query is routed to the semantic search stage, the system retrieves the top two most relevant chunks from the vector database, based on their similarity to the query. These chunks are subsequently reconverted to their original textual format.

This tool enables advanced log analysis by identifying relevant log sections based on their semantic meaning to the user's query, rather than just relying on keyword matching. By retrieving the top two most similar chunks, the tool provides contextually relevant results, ultimately enhancing log management efficacy and system troubleshooting capabilities.

\begin{figure}[H]
    \centering
    \includegraphics[width=0.8\linewidth]{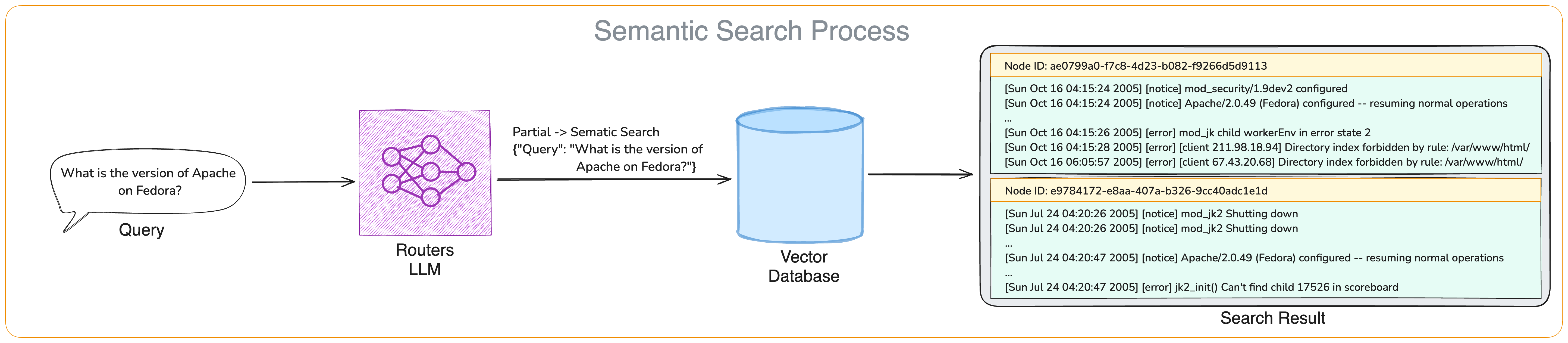}
    \caption{Semantic Search Process}
    \label{fig:semantic-search}
\end{figure}

For instance, as illustrated in Figure \ref{fig:semantic-search}, when a user queries "What is the Apache version on the Fedora" the semantic search tool retrieves the top two most similar log chunks, providing relevant information about the Apache version installed on the Fedora system.

\subsection{Generation}
\begin{figure}[H]
    \centering
    \includegraphics[width=0.8\linewidth]{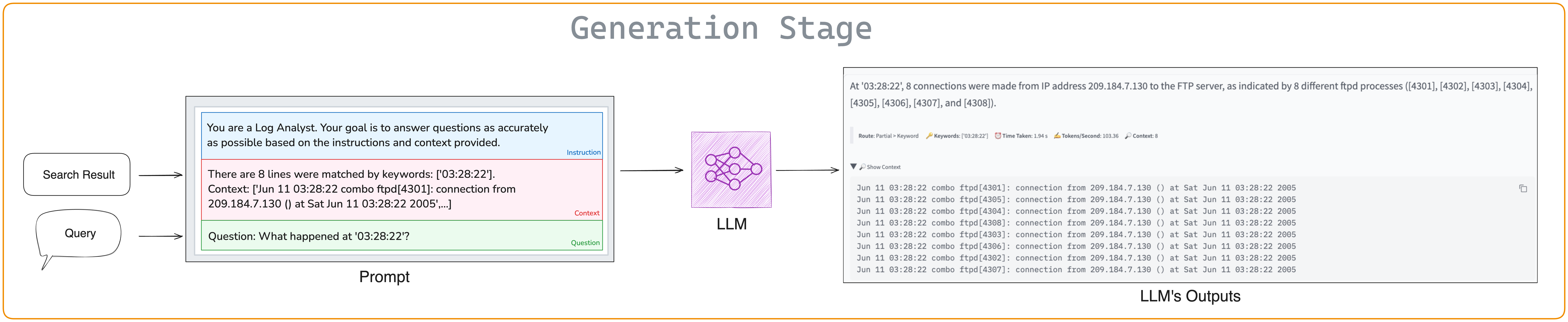}
    \caption{Generation Stage}
    \label{fig:generation}
\end{figure}

In the final stage, the LLM receives the appropriate prompt template determined by the router stage outcome, along with any available context from search results and the user's query. The complete set of prompt templates is available in appendices \ref{subsec:all_stage_prompt_template} -- \ref{subsec:event_prompt_template}. The LLM subsequently generates a response that answers the user's question while incorporating search results as supporting references, as shown in Figure \ref{fig:generation}.

\section{Experiments Setup}
\label{sec:experiments}

This section presents the experimental design and setup, devised to evaluate the effectiveness and robustness of LLMLogAnalyzer across multiple log analysis tasks on various datasets. The performance is comprehensively assessed through comparisons with other state-of-the-art candidates, using two distinct evaluation metrics.

\subsection{Baseline Models}

This study utilizes the open-source Llama 3 models \cite{Llama2024}, specifically the Llama-3-70B and Llama-3-8B variants, to assess the impact of model size on performance in log analysis tasks. In addition, three general-purpose LLM-based chatbots-ChatGPT (GPT-4o) \cite{ChatGPT2024}, ChatPDF \cite{ChatPDF2024}, and NotebookLM \cite{NotebookLM2024}-are included to benchmark against LLMLogAnalyzer. While these chatbots are not specialized for structured logs, their inclusion allows us to evaluate whether LLMLogAnalyzer provides measurable improvements over generic LLMs when applied to log analysis tasks. We acknowledge the limitations of using non-specialized tools as baselines and plan to extend our evaluation in future work by including domain-specific deep learning approaches such as LogPPT, PreLog, LogGPT, DivLog \cite{xu2024divlog}, and ULog \cite{huang2024ulog} for a more comprehensive comparison. 

\subsection{Log Dataset}

Four log datasets—Apache, Linux, macOS, and Windows—including both service and operating system traces, are selected from Loghub \cite{Zhu2023loghub} to conduct a proof-of-concept, given their widespread use in log analysis research. To evaluate scalability and performance on large-scale datasets, future benchmarking is planned using the full datasets from these sources, and even larger datasets, including BGL with 4 million entries, HDFS comprising more than 11 million rows of logs, and Thunderbird containing 16 million entries. Additionally, future work will incorporate logs from security-related sources (e.g., CERT and LANL), container environments (e.g., Kubernetes), and application-level systems (e.g., OpenStack).

\subsection{Log Analysis Tasks}
Seven log analysis tasks were then designed to assess those applications abilities comprehensively. These tasks include summarization, pattern extraction, log-based anomaly detection, root cause analysis, predictive failure analysis, log understanding and interpretation, and log filtering and searching. For each task category, multiple specific questions were systematically crafted to assess different aspects of application performance. Ground truth answers were established through comprehensive task-oriented manual analysis, where each of the seven log analysis tasks requires distinct analytical approaches. For summarization tasks, the analysis focuses on identifying key system events and overall behavior trends within the datasets. Pattern extraction requires systematic identification of recurring sequences and behavioral signatures through tracking event frequencies and temporal correlations. Log-based anomaly detection concentrated on establishing baseline normal behaviors and identifying deviations that indicate security threats or system malfunctions. Root cause analysis demands tracing causal relationships between events through chronological analysis and logical inference to connect symptoms with underlying causes. Predictive failure analysis focuses on identifying early warning indicators and patterns that precede system failures, recognizing subtle warning signs and escalating conditions. Log understanding and interpretation emphasize semantic comprehension of technical terminology, error codes, and contextual relationships within log entries. Log filtering and searching require developing precise search strategies and filtering criteria for efficient information retrieval.

\subsection{Evaluation Metrics}
To evaluate four different LLMs on seven log analysis tasks, the process begins with uploading a standardized log file to ensure consistency. This log file is then used to generate responses for each of the seven tasks in the four LLM-based applications. Once the responses are saved, the same tasks are entered into each model, and their outputs are compared and assessed against expected answers.

The model performance is assessed through multiple evaluation metrics. Cosine similarity quantifies the semantic alignment between generated responses and expected answers, while ROUGE-1 measures word-level overlap between generated outputs and reference content using precision, recall, and F1-score metrics, with particular emphasis on anomaly detection tasks. Future evaluations will incorporate top-k accuracy for root cause analysis applications.

These criteria effectively measure how well each model identifies and analyses relevant information within the log data, ensuring a thorough assessment of its capabilities.

\subsection{Implementation and Settings}

The experiments were conducted via invoking the Groq \cite{Groq2024} API of two variant Llama-3 models, specifically "llama-3.1-8b-instant" and "llama-3.1-70b-versatile". The Beijing Academy of Artificial Intelligence (BAAI) \cite{Baai2024} model, specifically "bge-base-en-v1.5", is utilized for embedding. A comprehensive evaluation is conducted, covering sixteen questions across seven categories of tasks. These tasks are queried on each LLM-based chatbot, using four different datasets. The evaluation metrics are calculated using the Natural Language Toolkit (NLTK) \cite{bird2009natural} library for cosine similarity and the ROUGE library for ROUGE-1 F1 score.

For log parsing, we utilize the same settings from Loghub 2.0 for the Drain algorithm, including log format, regex, st, and depth. For the Llama-3 models on Groq, the temperature is set to 0.7, which provides a suitable balance between response diversity and coherence. Chunking is employed with a chunk size of 1024 tokens. LLMLogAnalyzer is integrated with Gradio\cite{Gradio2024} to provide an intuitive and user-friendly web interface, facilitating easy interaction with the log analysis tool.

\section{Result}
\label{sec:result}

This section presents the findings of our experimental evaluation, providing insights into the performance of LLMLogAnalyzer in log analysis tasks. 

\subsection{RQ1: How does the LLMLogAnalyzer perform in log analysis tasks compared to other LLM-based applications?}

This research question investigates the performance of LLMLogAnalyzer in log analysis tasks, comparing it to other state-of-the-art LLM-based applications. The objective is to conduct a comprehensive evaluation of the proposed method's effectiveness.

To ensure a fair comparison, LLMLogAnalyzer is benchmarked against ChatGPT (GPT-4), ChatPDF, and NotebookLM using the same Loghub 2.0 datasets (e.g., Linux) for consistency. The evaluation encompasses seven distinct log analysis tasks, each with specifically defined questions to test robustness.
A comprehensive evaluation was conducted to assess LLMLogAnalyzer's robustness and effectiveness in log analysis tasks. The evaluation spanned seven tasks across four datasets, comparing LLMLogAnalyzer's performance to three other state-of-the-art LLM-based chatbots using cosine similarity and ROUGE-1 F1 score metrics.

\subsubsection{Effectiveness}

The effectiveness evaluation investigates LLMLogAnalyzer's generalization capability across diverse log data sources. The experimental setup utilized datasets of varying sizes to reflect the availability and characteristics of different log types: 20,000 rows for Apache, Linux, and Mac log datasets, and 5,000 rows for the Windows dataset. The results are presented in Tables \ref{tab:performance_comparison_apache}--\ref{tab:performance_comparison_windows}.

\begin{table}[H]
    \centering
    \caption{Performance Comparison in Apache}
    \resizebox{\textwidth}{!}{ 
    \begin{tabular}{@{}lcccccccccc@{}}
        \toprule
        & \multicolumn{2}{c}{\textbf{ChatGPT(GPT-4o)}} & \multicolumn{2}{c}{\textbf{NotebookLM}} & \multicolumn{2}{c}{\textbf{ChatPDF}} & \multicolumn{2}{c}{\textbf{LLMLogAnalyzer(Llama-3-70B)}} & \multicolumn{2}{c}{\textbf{LLMLogAnalyzer(Llama-3-8B)}} \\
        \cmidrule(lr){2-3} \cmidrule(lr){4-5} \cmidrule(lr){6-7} \cmidrule(lr){8-9} \cmidrule(lr){10-11}
        \textbf{Task} & Cosine Similarity & F1 & Cosine Similarity & F1 & Cosine Similarity & F1 & Cosine Similarity & F1 & Cosine Similarity & F1 \\
        \midrule
        Summarization & 0.26 & 0.33 & 0.2 & 0.31 & 0.22 & 0.26 & 0.41 & 0.51 & 0.36 & 0.42 \\
        Pattern Extraction & 0.12 & 0.23 & 0.08 & 0.18 & 0.07 & 0.21 & 0.38 & 0.39 & 0.05 & 0.1 \\
        Log-based Anomaly Detection & 0.11 & 0.23 & 0.14 & 0.25 & 0.19 & 0.26 & 0.49 & 0.53 & 0.19 & 0.26 \\
        Root Cause Analysis & 0.32 & 0.43 & 0.39 & 0.37 & 0.26 & 0.37 & 0.41 & 0.45 & 0.09 & 0.11 \\
        Predictive Failure Analysis & 0.18 & 0.36 & 0.19 & 0.29 & 0.27 & 0.4 & 0.59 & 0.57 & 0.2 & 0.25 \\
        Log Understanding and Interpretation & 0.47 & 0.28 & 0.44 & 0.2 & 0.47 & 0.22 & 0.55 & 0.42 & 0.51 & 0.32 \\
        Log Filtering and Searching & 0.25 & 0.29 & 0.1 & 0.08 & 0.17 & 0.1 & 0.46 & 0.44 & 0.25 & 0.28 \\
        \bottomrule
    \end{tabular}
    }
    \label{tab:performance_comparison_apache}
\end{table}

\begin{table}[h]
    \centering
    \caption{Performance Comparison in Linux}
    \resizebox{\textwidth}{!}{ 
    \begin{tabular}{@{}lcccccccccc@{}}
        \toprule
        & \multicolumn{2}{c}{\textbf{ChatGPT(GPT-4o)}} & \multicolumn{2}{c}{\textbf{NotebookLM}} & \multicolumn{2}{c}{\textbf{ChatPDF}} & \multicolumn{2}{c}{\textbf{LLMLogAnalyzer(Llama-3-70B)}} & \multicolumn{2}{c}{\textbf{LLMLogAnalyzer(Llama-3-8B)}} \\
        \cmidrule(lr){2-3} \cmidrule(lr){4-5} \cmidrule(lr){6-7} \cmidrule(lr){8-9} \cmidrule(lr){10-11}
        \textbf{Task} & Cosine Similarity & F1 & Cosine Similarity & F1 & Cosine Similarity & F1 & Cosine Similarity & F1 & Cosine Similarity & F1 \\
        \midrule
        Summarization & 0.18 & 0.21 & 0.13 & 0.2 & 0.11 & 0.17 & 0.56 & 0.45 & 0.28 & 0.37 \\
        Pattern Extraction & 0.15 & 0.3 & 0.1 & 0.24 & 0.2 & 0.29 & 0.41 & 0.48 & 0.33 & 0.41 \\
        Log-based Anomaly Detection & 0.13 & 0.23 & 0.12 & 0.22 & 0.19 & 0.28 & 0.43 & 0.47 & 0.08 & 0.11 \\
        Root Cause Analysis & 0.48 & 0.14 & 0.47 & 0.24 & 0.5 & 0.23 & 0.57 & 0.44 & 0.52 & 0.21 \\
        Predictive Failure Analysis & 0.21 & 0.29 & 0.21 & 0.2 & 0.35 & 0.26 & 0.54 & 0.58 & 0.21 & 0.22 \\
        Log Understanding and Interpretation & 0.61 & 0.58 & 0.33 & 0.31 & 0.43 & 0.39 & 0.47 & 0.49 & 0.34 & 0.33 \\
        Log Filtering and Searching & 0.26 & 0.24 & 0.1 & 0.09 & 0.09 & 0.05 & 0.4 & 0.42 & 0.28 & 0.25 \\
        \bottomrule
    \end{tabular}
    }
    \label{tab:performance_comparison_linux}
\end{table}

\begin{table}[h]
    \centering
    \caption{Performance Comparison in Mac}
    \resizebox{\textwidth}{!}{ 
    \begin{tabular}{@{}lcccccccccc@{}}
        \toprule
        & \multicolumn{2}{c}{\textbf{ChatGPT(GPT-4o)}} & \multicolumn{2}{c}{\textbf{NotebookLM}} & \multicolumn{2}{c}{\textbf{ChatPDF}} & \multicolumn{2}{c}{\textbf{LLMLogAnalyzer(Llama-3-70B)}} & \multicolumn{2}{c}{\textbf{LLMLogAnalyzer(Llama-3-8B)}} \\
        \cmidrule(lr){2-3} \cmidrule(lr){4-5} \cmidrule(lr){6-7} \cmidrule(lr){8-9} \cmidrule(lr){10-11}
        \textbf{Task} & Cosine Similarity & F1 & Cosine Similarity & F1 & Cosine Similarity & F1 & Cosine Similarity & F1 & Cosine Similarity & F1 \\
        \midrule
        Summarization & 0.17 & 0.23 & 0.25 & 0.24 & 0.23 & 0.28 & 0.78 & 0.73 & 0.33 & 0.29 \\
        Pattern Extraction & 0.23 & 0.32 & 0.22 & 0.30 & 0.25 & 0.31 & 0.40 & 0.33 & 0.03 & 0.15 \\
        Log-based Anomaly Detection & 0.18 & 0.17 & 0.15 & 0.17 & 0.21 & 0.20 & 0.30 & 0.29 & 0.06 & 0.14 \\
        Root Cause Analysis & 0.38 & 0.31 & 0.59 & 0.36 & 0.30 & 0.21 & 0.51 & 0.46 & 0.47 & 0.57 \\
        Predictive Failure Analysis & 0.20 & 0.11 & 0.18 & 0.07 & 0.21 & 0.10 & 0.45 & 0.32 & 0.15 & 0.12 \\
        Log Understanding and Interpretation & 0.32 & 0.12 & 0.35 & 0.13 & 0.38 & 0.12 & 0.39 & 0.32 & 0.42 & 0.20 \\
        Log Filtering and Searching & 0.12 & 0.11 & 0.12 & 0.10 & 0.21 & 0.13 & 0.45 & 0.41 & 0.27 & 0.15 \\
        \bottomrule
    \end{tabular}
    }
    \label{tab:performance_comparison_mac}
\end{table}

\begin{table}[h]
    \centering
    \caption{Performance Comparison in Windows}
    \resizebox{\textwidth}{!}{ 
    \begin{tabular}{@{}lcccccccccc@{}}
        \toprule
        & \multicolumn{2}{c}{\textbf{ChatGPT(GPT-4o)}} & \multicolumn{2}{c}{\textbf{NotebookLM}} & \multicolumn{2}{c}{\textbf{ChatPDF}} & \multicolumn{2}{c}{\textbf{LLMLogAnalyzer(Llama-3-70B)}} & \multicolumn{2}{c}{\textbf{LLMLogAnalyzer(Llama-3-8B)}} \\
        \cmidrule(lr){2-3} \cmidrule(lr){4-5} \cmidrule(lr){6-7} \cmidrule(lr){8-9} \cmidrule(lr){10-11}
        \textbf{Task} & Cosine Similarity & F1 & Cosine Similarity & F1 & Cosine Similarity & F1 & Cosine Similarity & F1 & Cosine Similarity & F1 \\
        \midrule
        Summarization & 0.31 & 0.38 & 0.31 & 0.42 & 0.17 & 0.28 & 0.4 & 0.46 & 0.29 & 0.36 \\
        Pattern Extraction & 0.1 & 0.23 & 0.18 & 0.31 & 0.11 & 0.21 & 0.45 & 0.44 & 0.18 & 0.17 \\
        Log-based Anomaly Detection & 0.26 & 0.39 & 0.28 & 0.39 & 0.17 & 0.35 & 0.29 & 0.4 & 0.23 & 0.25 \\
        Root Cause Analysis & 0.39 & 0.42 & 0.32 & 0.36 & 0.28 & 0.31 & 0.49 & 0.44 & 0.3 & 0.33 \\
        Predictive Failure Analysis & 0.3 & 0.39 & 0.3 & 0.33 & 0.21 & 0.39 & 0.34 & 0.39 & 0.38 & 0.41 \\
        Log Understanding and Interpretation & 0.41 & 0.26 & 0.41 & 0.16 & 0.46 & 0.17 & 0.46 & 0.42 & 0.43 & 0.31 \\
        Log Filtering and Searching & 0.17 & 0.07 & 0.12 & 0.03 & 0.1 & 0.07 & 0.41 & 0.4 & 0.06 & 0.04 \\
        \bottomrule
    \end{tabular}
    }
    \label{tab:performance_comparison_windows}
\end{table}

LLMLogAnalyzer consistently exhibits high scores across multiple datasets and tasks, although it doesn't always achieve the highest scores. Notable examples of LLMLogAnalyzer's top performance include its strong showing in the log understanding task with Apache log data, where LLMLogAnalyzer (Llama-3-70B) scored 0.55 in cosine similarity and 0.42 in ROUGE-1 F1. The system also performed well in the summarization task with Linux log data, achieving 0.56 in cosine similarity and 0.45 in ROUGE-1 F1, and demonstrated particularly impressive results in the summarization task with Mac log data, where LLMLogAnalyzer (Llama-3-70B) achieved 0.78 in cosine similarity and 0.73 in ROUGE-1 F1. However, in the log understanding task with Linux log data, ChatGPT (GPT-4o) achieved higher scores with 0.61 in cosine similarity and 0.58 in ROUGE-1 F1, compared to LLMLogAnalyzer (Llama-3-70B), which scored 0.47 in cosine similarity and 0.49 in ROUGE-1 F1.

Table \ref{tab:average_performance} presents a comparative analysis of the average performance of LLMLogAnalyzer and its counterparts across two evaluation metrics and four datasets.

\begin{table}[h]
    \centering
    \caption{Average Performance of LLMLogAnalyzer} 
    \resizebox{\textwidth}{!}{ 
    \begin{tabular}{@{}lcccccccccc@{}}
        \toprule
        & \multicolumn{2}{c}{\textbf{ChatGPT(GPT-4o)}} & \multicolumn{2}{c}{\textbf{NotebookLM}} & \multicolumn{2}{c}{\textbf{ChatPDF}} & \multicolumn{2}{c}{\textbf{LLMLogAnalyzer(Llama-3-70B)}} & \multicolumn{2}{c}{\textbf{LLMLogAnalyzer(Llama-3-8B)}} \\
        \cmidrule(lr){2-3} \cmidrule(lr){4-5} \cmidrule(lr){6-7} \cmidrule(lr){8-9} \cmidrule(lr){10-11}
        \textbf{Task} & Cosine Similarity & F1 & Cosine Similarity & F1 & Cosine Similarity & F1 & Cosine Similarity & F1 & Cosine Similarity & F1 \\
        \midrule
        Summarization & 0.26 & 0.33 & 0.20 & 0.31 & 0.22 & 0.26 & 0.41 & 0.51 & 0.36 & 0.42 \\
        Pattern Extraction & 0.12 & 0.23 & 0.08 & 0.18 & 0.07 & 0.21 & 0.38 & 0.39 & 0.05 & 0.10 \\
        Anomaly Detection & 0.11 & 0.23 & 0.14 & 0.25 & 0.19 & 0.26 & 0.49 & 0.53 & 0.19 & 0.26 \\
        Root Cause Analysis & 0.32 & 0.43 & 0.39 & 0.37 & 0.26 & 0.37 & 0.41 & 0.45 & 0.09 & 0.11 \\
        Predictive Failure Analysis & 0.18 & 0.36 & 0.19 & 0.29 & 0.27 & 0.40 & 0.59 & 0.57 & 0.20 & 0.25 \\
        Log Understanding and Interpretation & 0.47 & 0.28 & 0.44 & 0.20 & 0.47 & 0.22 & 0.55 & 0.42 & 0.51 & 0.32 \\
        Log Filtering and Searching & 0.25 & 0.29 & 0.10 & 0.08 & 0.17 & 0.10 & 0.46 & 0.44 & 0.25 & 0.28 \\
\bottomrule
    \end{tabular}
    }
    \label{tab:average_performance}
\end{table}

According to Table \ref{tab:average_performance}, the average performance of LLMLogAnalyzer (Llama-3-70B) across seven tasks is 0.45 in cosine similarity and 0.47 in ROUGE-1 F1. Notably, LLMLogAnalyzer outperforms its counterparts by significant margins, achieving a 43\% improvement over ChatGPT (GPT-4o) in cosine similarity and 39\% in ROUGE-1 F1. The system demonstrates even stronger performance compared to ChatPDF, with 47\% improvement in both cosine similarity and ROUGE-1 F1 metrics. Against NotebookLM, LLMLogAnalyzer shows 47\% improvement in cosine similarity and 48\% improvement in ROUGE-1 F1, highlighting its superior capabilities across different baseline comparisons.

On average, LLMLogAnalyzer surpasses ChatGPT, ChatPDF, and NotebookLM by 45\% in cosine similarity and 44\% in ROUGE-1 F1, demonstrating outstanding and consistent performance across various log analysis tasks.

In addition, LLMLogAnalyzer demonstrates outstanding performance in specific log analysis tasks, achieving notable scores across different categories. For log filtering and search tasks, the system achieves 0.61 in cosine similarity and 0.68 in ROUGE-1 F1. In pattern extraction, it records 0.63 in cosine similarity and 0.39 in ROUGE-1 F1. For summarization tasks, LLMLogAnalyzer obtains 0.56 in cosine similarity and 0.45 in ROUGE-1 F1.

\subsubsection{Robustness}

The robustness of LLMLogAnalyzer is assessed through a comprehensive comparison with baseline applications, using box plots to visualize the distribution of evaluation metrics across multiple datasets. To further validate the statistical significance of the results, we also plan to incorporate additional analyses, including variance, confidence intervals, error bars, significance testing, and p-values.

\begin{figure}[H]
    \centering
    \includegraphics[width=0.8\linewidth]{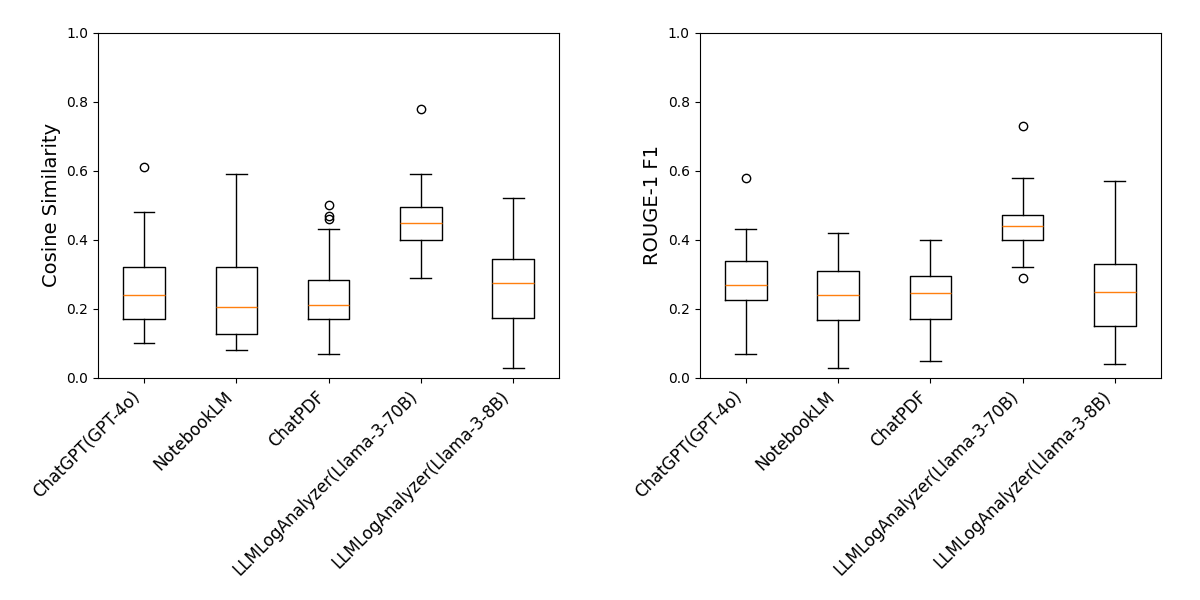}
    \caption{Robustness comparison}
    \label{fig:robustness}
\end{figure}

As illustrated in Figure \ref{fig:robustness}, LLMLogAnalyzer (Llama-3-70B) shows consistent robustness across multiple datasets, outperforming the others in both metrics. Specifically, LLMLogAnalyzer (Llama-3-70B) achieves higher median values of 0.45 for cosine similarity and 0.44 for ROUGE-1 F1, reflecting consistent high-quality performance. As evidenced by shorter box lengths in the box plot, this robustness is further characterized by a narrower Interquartile Range (IQR) - 34\% narrower for cosine similarity and 93\% narrower for ROUGE-1 F1 - indicating significantly lower variability. Furthermore, LLMLogAnalyzer (Llama-3-70B) exhibits remarkably low outlier rates of 3.6\% for cosine similarity and 7\% for ROUGE-1 F1, demonstrating exceptional reliability and stability.  

In comparison, ChatGPT(GPT-4o) performs consistently with moderate scores in both metrics, showing stable results and a small interquartile range (IQR). NotebookLM and ChatPDF underperform, with lower median scores and increased variability. On the other hand, LLMLogAnalyzer (Llama-3-70B) leads with the highest median scores and narrow IQRs, whereas LLMLogAnalyzer (Llama-3-8B) consistently shows the lowest median values and greatest fluctuation.

LLMLogAnalyzer's effectiveness and robustness are conclusively demonstrated through its consistent performance across a range of log datasets, making it a reliable tool for log analysis applications.

\subsection{RQ2: How does model size affect the performance of LLMLogAnalyzer in log analyzing tasks?}

This research question investigates the strengths and weaknesses of two LLMLogAnalyzer variants, Llama-3-70B and Llama-3-8B. The Llama-3-70B model provides more powerful reasoning ability, which is advantageous for log analysis tasks. In contrast, the Llama-3-8B model offers a more lightweight and efficient alternative. By comparing their performance across specific tasks, this analysis identifies task-specific strengths and weaknesses, uncovers performance trade-offs, and informs the selection of the appropriate variant based on task requirements.

The results, as shown in Tables \ref{tab:performance_comparison_apache}--\ref{tab:performance_comparison_windows}, indicate that the LLMLogAnalyzer (Llama-3-70B) model consistently outperforms the LLMLogAnalyzer (Llama-3-8B) model across almost all tasks and systems. This suggests that the larger 70B parameter model possesses a greater capacity for understanding and interpreting log data.

The LLMLogAnalyzer (Llama-3-70B) variant achieves significantly higher Cosine Similarity and ROUGE-1 F1 scores than the LLMLogAnalyzer (Llama-3-8B) variant in the majority of tasks and across all systems. This superior performance is evident in tasks like Summarization on Linux and Mac systems, where the LLMLogAnalyzer (Llama-3-70B) model scores 0.56 and 0.78 in Cosine Similarity, respectively, compared to the LLMLogAnalyzer (Llama-3-8B) model's scores of 0.28 and 0.33.

The LLMLogAnalyzer (Llama-3-70B) variant displays a notable performance advantage in tasks requiring more nuanced comprehension and interpretation of log data, such as root cause analysis and log understanding and interpretation. For example, in the log understanding and interpretation task on the Apache system, the LLMLogAnalyzer (Llama-3-70B) model achieves a cosine similarity score of 0.55, while the LLMLogAnalyzer (Llama-3-8B) model scores 0.51. This difference suggests that the larger model size contributes to a better understanding of the context and meaning within complex log data.

The lower scores of the LLMLogAnalyzer (Llama-3-8B) variant in comparison to the LLMLogAnalyzer (Llama-3-70B) variant across almost all evaluated scenarios indicate potential limitations in its ability to effectively process and analyze log data. This suggests that the smaller model size might restrict its capacity to learn and represent complex patterns in log data effectively.
        
The performance difference between the two variants widens in tasks requiring greater contextual understanding and interpretation. This suggests that the smaller model, LLMLogAnalyzer (Llama-3-8B), may struggle to effectively handle intricate log analysis scenarios demanding a deeper understanding of relationships and patterns within the data.

Notwithstanding these findings, the smaller model demonstrates comparable performance in pattern extraction and log understanding, and interpretation. This suggests that model size is a critical factor influencing performance, but smaller models may remain viable for specific log analysis tasks.


\section{Discussion}
\label{sec:discussion}

The results demonstrate that the larger LLMLogAnalyzer model is more effective and robust in handling multiple datasets and diverse log analysis tasks, particularly in reasoning. This remarkable performance can be attributed to the synergistic integration of ML and LLMs, which enables the chatbot to accurately identify, analyze, and interpret log data. Instead of directly answering user queries. LLMLogAnalyzer actively analyzes the query, apply a proper search tool to enhance accuracy, and provides response with references. 

LLMLogAnalyzer mitigates the contextual limitations of LLMs in analyzing the structured logs by integrating them with ML algorithms. It offers a comprehensive framework that eliminates the need for expertise in log analysis, enabling both log experts and non-technical users to perform log analysis with ease. Additionally, it offers effortless integration, requiring no training or fine-tuning processes. The solution is flexible, supporting various log categories and enabling users to integrate their models for optimized performance and enhanced data security with open-source models hosted on-premises.

This research contributes to the exploration of the applicability of such approaches in industrial settings, with potential future applications. However, this study also identifies some limitations that require further attention. Addressing these limitations will be critical to realizing the full potential of the LLMLogAnalyzer. Future research can build on this study and explore new ways to improve the performance, applicability, and scalability of LLMLogAnalyzer. The current study was conducted as a proof of concept, and while the framework shows promise, its performance at real-time, enterprise-level log data volumes remains to be validated. Such enterprise-scale deployments would involve processing terabytes of log data daily, requiring additional optimization of computational resources, response times, and perhaps the framework's architecture. 

Future work on LLMLogAnalyzer will focus on several key areas, including integrating database capabilities and Text-to-SQL functionality for more efficient log analysis, developing an interactive interface, and incorporating ReAct (Reason + Act) to enhance LLM performance. Furthermore, investigation of log parsing algorithm accuracy and its impact on LLM performance represents important future work, as inaccurate parsing may lead to flawed analysis results across all seven task categories. Additional enhancements will include conversation history and multi-user support to enable collaborative log analysis. The study also plans to benchmark LLMLogAnalyzer against domain-specific deep learning solutions, such as LogPPT, PreLog, LogGPT, DivLog, and ULog, and evaluate its scalability using large datasets like HDFS-2.8 (11M lines/24h), BGL (4.7M), ThunderBird (26M), and BlueGene/L (4.2B).


\section{Conclusion}
\label{sec:conclusion}
Building upon existing research, this study further explores the capabilities of LLMs for log analysis. However, LLMs' input token limits hinder their direct application to large log datasets. LLMLogAnalyzer is developed to address existing limitations by providing a cost-effective and scalable LLM-based log analysis framework. Leveraging ML algorithm, LLMLogAnalyzer efficiently processes raw log data into structured log data. Its modular design enables dynamic and precise processing of user queries, with three additional search tools further enhancing the accuracy of the response. Notably, LLMLogAnalyzer outperformed other baseline applications in most log analysis tasks, with the LLMLogAnalyzer (Llama-3-70B) variant emerging as the top performer. These findings establish a solid foundation for future research, focusing on optimizing LLM-based applications for log analysis and unlocking their full potential on large-scale log files.

\newpage
\section*{Funding}
No funding was received for this study.

\section*{Competing Interests}
The authors declare no competing interests.

\section*{Data Availability Statement}
The log datasets used in this study are from Loghub-2.0, which is publicly available at: \doi{10.5281/zenodo.8275860}. The processed results and source code developed for this study are publicly available at \doi{10.5281/zenodo.15798461} under the Creative Commons Attribution (CC-BY) license.

\newpage
 \bibliographystyle{elsarticle-num} 
 \bibliography{cas-refs}

\newpage
\appendix

\section{Prompt Templates}
\label{sec:appendix}

\subsection{System Prompt Template}
\label{subsec:system_prompt_template}

\begin{lstlisting}
<|begin_of_text|><|start_header_id|>system<|end_header_id|> You are a Log Analyst. Your goal is to answer questions as accurately as possible based on the instructions and context provided.<|eot_id|><|start_header_id|>user<|end_header_id|> {query_str}<|eot_id|><|start_header_id|>assistant<|end_header_id|>
\end{lstlisting}

\subsection{Log Recognizer Prompt Template}
\label{subsec:log_recognizer_prompt_template}

\begin{lstlisting}
You are a log expert tasked with categorizing a provided log line. The categories are: {categories}.
Categorize the provided log lines into one of the categories:{categories}, assuming those logs share a single category.
Return a JSON object with a single key "category" without any preamble, special characters, or explanation.
Log:\n{log}
\end{lstlisting}

\subsection{Routers Level 1 Prompt Template}
\label{subsec:router_level_1_prompt_template}

\begin{lstlisting}
You are an expert at routing user questions to the 'all', 'partial', or 'general' stage.
To answer the user's question, you must analyze how the question relates to the log content 
and decide whether to retrieve all, partial, or no logs from the log file.
- Use 'all' for questions requiring the entire log file to answer. 
- Use 'partial' for questions requiring only a specific part or chunk of the log file to answer. 
- Use 'general' for questions that can be answered without needing to retrieve any logs. 
Return a JSON as plaintext with a single key 'choice' based on the question without any preamble, special characters, or explanation.
Question to route: {question}
\end{lstlisting}

\subsection{Routers Level 2 Prompt Template}
\label{subsec:router_level_2_prompt_template}

\begin{lstlisting}
You are an expert at routing user questions to the 'keyword', 'event', or 'se' stage.
Use 'keyword' if the question requires using a search tool to find relevant information. If the question does not have clear keywords, try using 'semantic' instead.
- Use 'event' if the question is related to a specific event or log template in the log file.
- Use 'semantic' if the question asks for specific information that can be retrieved from a vector database.
Return a JSON as plaintext with a single key 'choice' based on the question without any preamble, special characters, or explanation. If 'keyword' is chosen, also return 'keywords' as a list.  If 'event' is chosen, also return 'events' as a list. 
Question to route: {question}
\end{lstlisting}

\subsection{'All Event' Stage Prompt Template}
\label{subsec:all_stage_prompt_template}

\begin{lstlisting}
Context information is below.
Log file name {log_file_name}
All the events in this log file, which is a CSV file, contain three columns: EventId,EventTemplate,Occurrences
{templates[1:]}
The first line of the log file: {log_lines[0].strip()}
The Last line of the log file: {log_lines[-1].strip()}
The log file contains {len(log_lines)} lines
There are {len(templates)} log events in the log file.
The log period can be indicated in the first and last lines of the log file.
Given the context information, answer the query.
Query: {question}
Answer:
\end{lstlisting}

\subsection{‘Partial-Retrieve’ Stage Prompt Template}
\label{subsec:retrieve_prompt_template}

This prompt remains the same as the LlamaIndex \cite{llamaindex2023} Prompt Engineering for RAG.

\begin{lstlisting}
 Context information is below.
---------------------
{context_str}
---------------------
Given the context information and not prior knowledge, answer the query.
Query: {query_str}
Answer:
\end{lstlisting}

\subsection{‘Partial-Search’ Stage Prompt Template}
\label{subsec:search_prompt_template}

\begin{lstlisting}
There are {len(search_result)} lines were matched by keywords: {keywords}.
The context is too long, and it has been trimmed to {MAX_LINES} lines.
Focus on the total number if the user asks a question about how many.
Context: {search_result_max if search_result_modified else search_result}
Question: {question}.
\end{lstlisting}

\subsection{‘Partial-Event’ Stage Prompt Template}
\label{subsec:event_prompt_template}

\begin{lstlisting}
There are {len(search_result)} lines were matched by events: {events}.
The context is too long, and it has been trimmed to {MAX_LINES} lines.
All relevant events are: {filtered_df.values.tolist()}
Focus on the total number if the user asks a question about how many.
Context: {search_result_max if search_result_modified else search_result}
Question: {question}.
\end{lstlisting}






\end{document}

%% file: algorithm.tex
\begin{algorithm}[!ht]
\caption{LLMLogAnalyzer Process}
\label{alg:llmloganalyzer}
\scriptsize 
\begin{algorithmic}[1]

\STATE \textbf{Input:} \\
    \quad $L$: Log file \\
    \quad $Q$: User query
\STATE \textbf{Output:} \\
    \quad $R$: LLM's response to a user query

\STATE

\STATE \textbf{Indexing}  
\STATE \quad $L_{\text{chunks}} \gets$ chunking($L$) 
\STATE \quad $L_{\text{vectors}} \gets$ embedding($L_{\text{chunks}}$)

\STATE \textbf{Parsing}
\STATE \quad $L_{\text{type}} \gets$ identifyLogType($L$) 
\STATE \quad $L_{\text{structured}} \gets$ parseLogWithDrainClustering($L$, $L_{\text{type}}$)
\STATE \quad $L_{\text{events}} \gets$ retrieveEvents($L$, $L_{\text{structured}}$)

\STATE

\STATE \textbf{Query:}
\STATE \quad $stage \gets$ routeQueryL1($Q$)

\STATE

\STATE \textbf{Generation:}

\IF{$stage = \textbf{All}$}
    \STATE \quad $R \gets$ llmProcess($Q$, $L_{\text{events}}$)
\ELSIF{$stage = \textbf{Partial}$}
    \STATE \quad $searchType \gets$ routeQueryL2($Q$) 
    \STATE \quad $results \gets \emptyset$
    \IF{$searchType = \textbf{Keyword}$}
        \STATE \quad $keywords \gets$ getKeywords($Q$) 
        \STATE \quad $results \gets$ searchLogWithKeywords($L_{\text{chunks}}$, $keywords$)
    \ELSIF{$searchType = \textbf{Event}$}
        \STATE \quad $eventIds \gets$ getEventIDs($Q$) 
        \STATE \quad $results \gets$searchLogWithEvents($L_{\text{events}}$, $eventIds$)
    \ELSE
        \STATE \quad $results \gets$ semanticSearch($L_{\text{vectors}}$, $Q$)
    \ENDIF
    \STATE $R \gets$ llmProcess($Q$, $results$)
\ELSE
    \STATE \quad $R \gets$llmProcess($Q$)
\STATE \quad \ENDIF

\STATE

\RETURN $R$

\end{algorithmic}
\end{algorithm}